\documentclass[11pt]{article}

\usepackage{acl}

\usepackage{times}
\usepackage{latexsym}
\usepackage{xspace}
\usepackage{amsmath}
\usepackage{times}  
\usepackage{helvet}  
\usepackage{courier}  
\usepackage{booktabs}       
\usepackage{amsfonts}       
\usepackage{nicefrac}       
\usepackage{microtype}      
\usepackage{xcolor}         
\usepackage{caption} 
\usepackage[T1]{fontenc}

\usepackage[utf8]{inputenc}
\usepackage[most]{tcolorbox}

\usepackage{algorithm}
\usepackage{algorithmic}
\usepackage{array}
\usepackage{amsthm}

\usepackage{float}
\newcommand{\methodFont}{\texttt}
\usepackage{enumitem}

\newcommand{\ours}{\methodFont{CFA}\xspace}
\usepackage{microtype}

\usepackage{inconsolata}

\usepackage{graphicx}

\title{Conformal Feedback Alignment: Quantifying Answer-Level Reliability for Robust LLM Alignment}

\author{
  Tiejin Chen \\
  Arizona State University \\
  \texttt{tchen169@asu.edu}
  \And
  Xiaoou Liu \\
  Arizona State University \\
  \texttt{xiaoouli@asu.edu}
  \And
  Vishnu Nandam \\
  Arizona State University \\
  \texttt{vnandam@asu.edu}
  \AND
  Kuan-Ru Liou \\
  Arizona State University \\
  \texttt{kliou@asu.edu}
  \And
  Hua Wei \\
  Arizona State University \\
  \texttt{hua.wei@asu.edu}
}

\begin{document}
\maketitle
\begin{abstract}
Preference-based alignment like Reinforcement Learning from Human Feedback (RLHF) learns from pairwise preferences, yet the labels are often noisy and inconsistent. Existing uncertainty-aware approaches weight preferences, but ignore a more fundamental factor: the reliability of the \emph{answers} being compared. To address the problem, we propose Conformal Feedback Alignment (CFA), a framework that grounds preference weighting in the statistical guarantees of Conformal Prediction (CP). CFA quantifies answer-level reliability by constructing conformal prediction sets with controllable coverage and aggregates these reliabilities into principled weights for both DPO- and PPO-style training. Experiments across different datasets show that CFA improves alignment robustness and data efficiency, highlighting that modeling \emph{answer-side} uncertainty complements preference-level weighting and yields more robust, data-efficient alignment. Codes are provided \href{https://github.com/tiejin98/Conformal-Feedback-Alignment}{here}.
\end{abstract}

\section{Introduction}

Large Language Models (LLMs) have achieved remarkable capabilities~\cite{anil2023palm,achiam2023gpt,touvron2023llama}, largely driven by alignment techniques that fine-tune them on human preferences, such as Reinforcement Learning from Human Feedback (RLHF)~\citep{ouyang2022training,christiano2017deep} and Direct Preference Optimization (DPO)~\citep{rafailov2023direct}. To overcome the cost and scalability limitations of human annotation, the field is increasingly adopting Reinforcement Learning from AI Feedback (RLAIF), where preference data is generated by capable AI evaluators~\cite{yu2024rlaif,lee2023rlaif,li2024hrlaif}. While this method is powerful, it generates new challenges: preference labels can be noisy and inconsistent, which creates a fundamental noisy label problem that can degrade the quality and robustness of the final aligned model~\cite{banerjee2024towards,wang2024uncertainty}.

Recent studies have explored uncertainty-aware preference alignment, aiming to handle feedback by estimating uncertainty at the preference level. For example, \citet{banerjee2024towards} estimates reward-model uncertainty through ensemble variance, \citet{lodkaewimportance} applies weights to DPO losses, \citet{xu2024uncertainty} introduces a Bayesian formulation that models preference uncertainty within a risk-sensitive policy framework, and WPO~\cite{zhou2024wpo} reweights pairs by likelihood to mitigate off-policy drift.  All these approaches reduce the effect of noisy preference pairs by modeling uncertainty on preferences, i.e., how reliable the preference relationship is.

However, these methods overlook a more fundamental dimension of uncertainty: the intrinsic reliability of the individual answers that constitute the preference pair. A preference judgment can only be as trustworthy as the responses it compares. When both answers are of low confidence and quality, the preference carries little meaningful information, regardless of how we model uncertainty for the preference itself. This dimension of answer-level reliability represents a critical blind spot in current research, as it addresses the quality of the data at its very source. Therefore, we argue that each model-generated answer carries its own reliability, which directly contributes to the uncertainty observed in preference comparisons.


To address this distinct dimension of answer-level reliability, we introduce Conformal Feedback Alignment (\ours), a novel framework that grounds preference weighting in the statistical guarantees of Conformal Prediction (CP)~\cite{quach2023conformal,su2024api}. Instead of focusing on the downstream learning process, CFA directly assesses the quality of the answer itself. The core of our method is to first use CP to construct statistically valid prediction sets for responses with a pre-defined and controllable coverage. A higher-coverage prediction set is considered more reliable than a lower-coverage set. We then introduce a set-wise uncertainty aggregation function that translates the set memberships of responses into a single, principled weight that quantifies the trustworthiness of the preference label. This weight is integrated into both PPO-style and DPO-style alignment, forcing the model to prioritize judgments built upon high-confidence answers. Overall, our primary contributions are:

\begin{itemize}[leftmargin=*,noitemsep]
    \item We identify and address a critical, under-explored dimension in uncertainty-aware alignment: the intrinsic reliability of individual answers, which is distinct from uncertainty measured at the preference level.
    \item To the best of our knowledge, \ours is the first to construct  statistically valid prediction sets with Conformal Prediction for AI feedback, which is flexible for both black- and white-box settings as well as DPO- and PPO-style model training.
    \item Through comprehensive experiments, we demonstrate that by focusing on answer reliability, CFA improves alignment performance and data efficiency, outperforming standard baselines and offering comparable performance over methods that address preference-level uncertainty.
\end{itemize}

\section{Related Works}

\subsection{LLM Alignment}
LLMs have achieved strong performance on a wide range of tasks~\citep{chen2025privacy,chen2025protecting,satheesh2025cmalc,da2024prompt,da2024open,yao2025comal,da2025survey,da2025ge,da2025flans,da2025deepshade,yao2025instructional}. A key driver of this success is Reinforcement Learning from Human Feedback (RLHF)~\citep{christiano2017deep}, which aligns model outputs with human preferences. 
RLHF is traditionally implemented as a \textbf{reward-model-based method}, typically using Proximal Policy Optimization (PPO)~\citep{schulman2017proximal}. This approach requires training both a reward model and a policy model. To simplify this process, several alternatives have been proposed. REINFORCE++~\citep{hu2025reinforce++} eliminate the need for a critic model.

Beyond reward-based methods, \textbf{reward-model-free method} Direct Preference Optimization (DPO)~\citep{rafailov2023direct} directly optimizes LLM parameters based on pairwise preferences, bypassing reward model training. Extensions such as IPO~\citep{azar2024general}, $\alpha$-DPO~\citep{wu2024alpha}, CPO~\citep{xu2024contrastive}, TPO~\citep{saeidi2024triple}, and KTO~\citep{ethayarajh2024kto} improve DPO's stability, adaptiveness, or data efficiency. 

To address the cost of human labels, recent work has explored Reinforcement Learning from AI Feedback (RLAIF)~\citep{bai2022constitutional}. AI-generated preference data can reduce labeling cost and even improve alignment performance~\citep{lee2023rlaif,li2024hrlaif,williams2024multi}, while AI-generated preferences might be more noisy. To address this problem, several methods that provide weight to the preference pair are proposed~\cite{ye2025robust,lin2023generating,xu2024uncertainty,wang2024uncertainty}. However, all of them are focusing on the reliability of the preference without considering the reliability of responses.

\subsection{Uncertainty Quantification}

Uncertainty quantification (UQ) has been studied in LLMs~\citep{lin2023generating,chen2025uncertainty,liu2025uncertainty,chen2025uncertainty,da2024llm,liu2025mcqa}, but most prior work focuses on token-level or output-level generation uncertainty~\cite{liu2025uncertainty}. For example,~\citet{kadavath2022language} and \citet{band2024linguistic} show that LLMs often produce poorly calibrated self-assessments. 
Conformal Prediction (CP)~\citep{shafer2008tutorial} provides a non-parametric, distribution-free approach for constructing confidence sets with formal guarantees. Recent extensions adapt CP to LLMs in both white-box settings~\citep{quach2023conformal} and black-box API scenarios~\citep{su2024api}. However, existing UQ, including CP methods, are rarely integrated into preference learning or alignment pipelines. 

\section{Preliminaries}
\label{sec:prelim}

\begin{figure*}[t!]
\centering
\includegraphics[width=0.85\textwidth]{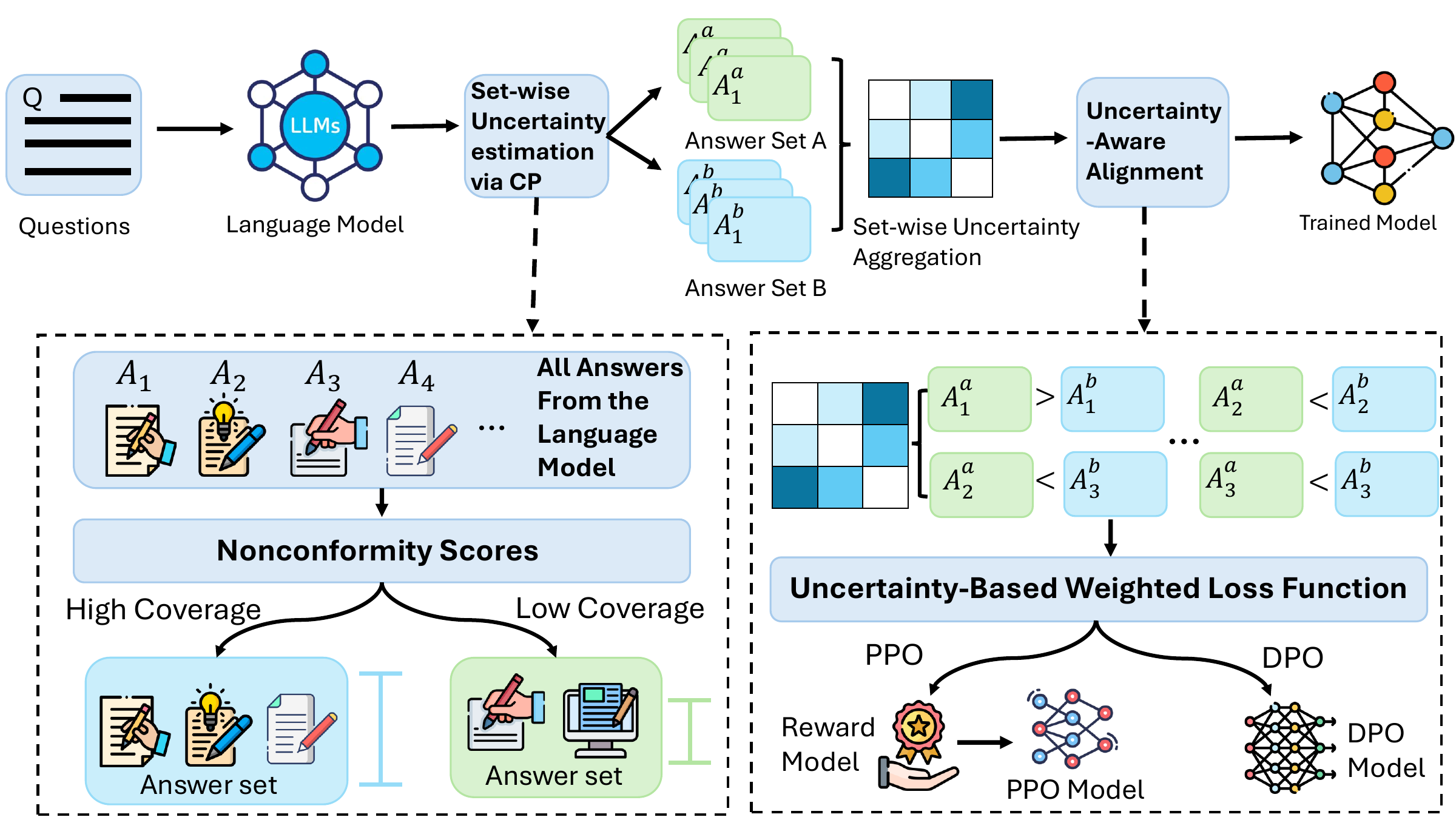}
\caption{The pipeline of our method. In our paper, we use conformal prediction, which can be applied for both black-box and white-box settings, to estimate the uncertainty and then use a weighted loss to conduct alignment.}
\label{fig:pipeline}
\vspace{-5mm}
\end{figure*}

\subsection{Preference-Based Alignment}

Reinforcement learning from human feedback (RLHF) and its variants optimize large language models (LLMs) by learning from preference comparisons. Given a prompt $x$, two outputs $y^+$ (preferred) and $y^-$ (dispreferred), the dataset is denoted as: $
\mathcal{D}_{\text{pref}} = {(x, y^+, y^-)}.
$
Two main approaches exist for preference-based alignment:

\paragraph{Reward-Model-Based Methods (PPO-style).}

RLHF typically trains a reward model $r_\phi(x, y)$ using pairwise comparisons. The standard loss is:

\begin{equation}
\small
\begin{aligned}
&\mathcal{L}_{\text{RM-Orig}}(\phi) \\
&= -\mathbb{E}_{(x, y^+, y^-) \sim \mathcal{D}_{\text{pref}}}
    \log \sigma \left( r_\phi(x, y^+) - r_\phi(x, y^-) \right),
\end{aligned}
\label{eq:rm-orig}
\end{equation}
where $\sigma(\cdot)$ is the sigmoid function.

After training $r_\phi$, a policy $\pi_\theta$ is optimized using Proximal Policy Optimization (PPO)~\citep{schulman2017proximal} with $r_\phi(x, y)$ as the reward.

\paragraph{Reward-Free Methods (DPO-style).}

Direct Preference Optimization (DPO)~\citep{rafailov2023direct} directly optimizes the policy $\pi_\theta$ without training a reward model. The loss is:

\begin{equation}
\mathcal{L}_{\text{DPO-Orig}}(\theta) = -\mathbb{E}_{(x, y^+, y^-) \sim \mathcal{D}_{\text{pref}}} \log \sigma\left( \beta \cdot \Delta_{\theta,\text{ref}} \right),
\label{eq:dpo-orig}
\end{equation}
where $\beta$ is a temperature parameter, and:

\begin{equation}
\Delta_{\theta, \text{ref}} = \log \frac{ \pi_\theta(y^+|x) }{ \pi_{\text{ref}}(y^+|x) } - \log \frac{ \pi_\theta(y^-|x) }{ \pi_{\text{ref}}(y^-|x) }.
\end{equation}

Here, $\pi_{\text{ref}}$ is a frozen reference policy, typically the supervised fine-tuned model.

\subsection{Conformal Prediction}

\label{sec:pre-cp}

Uncertainty aims to provide reliable confidence in prediction results~\citep{kadavath2022language,band2024linguistic}. Conformal Prediction (CP) is a distribution-free method that provides statistically valid uncertainty estimates.
The core idea of CP is to compute a nonconformity score $s(x, y)$ for each output $y$ given a prompt $x$. This score quantifies how ``unusual" or ``atypical" the output is. Lower scores indicate that the output is typical or confident, while higher scores suggest uncertainty.

Given a calibration set with known outputs, nonconformity scores are computed for each example, and a quantile threshold $q_\alpha$ is determined corresponding to the desired coverage level $1-\alpha$.
The conformal set is:
\begin{equation}
\mathcal{C}_{1-\alpha}(x) = \{ y \mid s(x, y) \leq q_\alpha \}.
\end{equation}
This set is guaranteed to include a correct output with probability at least $1-\alpha$.

\vspace{1mm}
\noindent$\bullet$~\textbf{White-Box CP:} 
In the white-box setting, token-level log-probabilities are available. Therefore, the nonconformity score can be defined as the negative log-likelihood of the output~\citep{quach2023conformal}:
\begin{equation}
s(x, y) = -\log p_\theta(y|x).
\end{equation}
This choice of scoring function has a straightforward interpretation. If the model assigns high probability to $y$ given $x$, then $s(x, y)$ will be small, indicating that the model is confident in this output. Conversely, if the model assigns low probability to $y$, the score will be large, reflecting higher uncertainty.


\noindent$\bullet$~\textbf{Black-Box CP:} 
In the black-box setting, where token-level probabilities are unavailable, the nonconformity score is estimated from the multiple model-generated samples and is given by~\citep{su2024api}:
\begin{equation}
s(x, y) = -\mathrm{Freq}(y) + \lambda_1 \cdot \mathrm{NE}(x) - \lambda_2 \cdot \mathrm{Sim}(y, y_{\text{top}}),
\end{equation}
%
where $\mathrm{Freq}(y)$ is the count of $y$ among the sampled responses for prompt $x$,
$\mathrm{NE}(x)$ is the normalized entropy of the sampled response distribution,
$\mathrm{Sim}(y, y_{\text{top}})$ is the similarity to the most frequent sample $y_{\text{top}}$,
and $\lambda_1,\lambda_2$ are weighting coefficients.
It suggests that responses that occur more frequently and are more similar to the most frequent sample yield lower nonconformity scores, indicating higher confidence and lower uncertainty.



\section{Method}
This section presents Conformal Feedback Alignment (\ours). Our goal is to improve the robustness of alignment using the answer reliability from CP.

\subsection{Overview}


Figure~\ref{fig:pipeline} presents an overview of our framework, which has two components: (1) estimating the uncertainty of AI-generated answers, which is considered as the answer reliability, using CP and (2) incorporating this uncertainty into both reward-model-based (e.g., PPO) and reward-free (e.g., DPO) alignment. For each preference pair, \ours produces an uncertainty score $u \in [0,1]$ from the answer reliability, which is used to weight the learning signal during policy optimization, prioritizing reliable feedback upon high-reliability answers while reducing the effect of comparisons for less reliable answers. The next sections describe set-wise uncertainty estimation (Section~\ref{sec:cp-uq}) and its integration into alignment (Section~\ref{sec:upl}).

\subsection{Set-wise Uncertainty Estimation via CP}
\label{sec:cp-uq}

To account for the reliability of responses, a principled approach to uncertainty estimation is essential. However, LLMs are known to produce uncalibrated uncertainty scores~\citep{band2024linguistic}. We address this by employing CP (Section~\ref{sec:pre-cp}) to generate statistically valid prediction sets. After employing CP, we can obtain two sets A and B with different coverage $\alpha$. The properties of these sets form the basis for our \textbf{Set-wise Uncertainty Aggregation} method, which translates answer-level reliability into a weight for the preference pair.


For a given preference pair $(y^+, y^-)$, we define a set-wise uncertainty score $u(x, y^+, y^-)$ based on the conformal sets to which the two responses belong. If both $y^+$ and $y^-$ belong to the set A, we use the corresponding quantile $q_a$ as the confidence. If they both belong to set B, we use $q_b$. If the two outputs belong to different sets, we take the average of the two quantiles. Formally, we define:
\begin{equation}
\small
\label{eq:uncertainty}
u(x, y^+, y^-) =
\begin{cases}
q_a & \text{if both } y^+, y^- \in \text{set A}, \\
q_b & \text{if both } y^+, y^- \in \text{set B}, \\
\frac{q_a + q_b}{2} & \text{if } y^+, y^- \text{ are from different sets}.
\end{cases}
\end{equation}

This design reflects the overall reliability of the comparison based on the reliability of the answers. When the outputs come from uncertain sets or the sets are mismatched, the confidence is reduced because the answers are not reliable. This set-wise approach allows us to calibrate the preference signal based on the uncertainty in the model's predictions.

\subsection{Uncertainty-Aware Alignment}
\label{sec:upl}




Standard alignment objectives, as used in PPO and DPO (Section~\ref{sec:prelim}), assume all comparisons are equally reliable. To address this limitation, we use uncertainty scores to adjust the training process. We introduce an uncertainty-weighted loss function, where each preference pair's contribution is scaled by its estimated reliability. This ensures that more trustworthy comparisons have a greater influence on alignment.

\paragraph{Reward-Model-Based Optimization (PPO-Style)}
In the standard PPO-style approach, a reward model is trained by optimizing the preference loss shown in Eq.~\ref{eq:rm-orig}. We adapt this objective to be uncertainty-aware by introducing an uncertainty weight $u$ for each comparison: 

\begin{equation}
\small
\label{eq:uppo}
\begin{aligned}
&\mathcal{L}_{\text{RM-Uncert}}(\phi) \\ 
& = -\mathbb{E}_{(x, y^+, y^-, u)} u \cdot \log \sigma \left( r_\phi(x, y^+) - r_\phi(x, y^-) \right)
\end{aligned}
\end{equation}



\paragraph{Reward-Free Optimization (DPO-Style)}
For reward-free learning, we build on the DPO objective defined in Eq.~\ref{eq:dpo-orig}. DPO optimizes the policy directly by maximizing the log odds of the preferred output relative to the reference policy. To account for uncertainty, we introduce the uncertainty weight $u$ from Set-wise Uncertainty Aggregation into this objective, multiplying the log-sigmoid term by $u$:
\begin{equation}
\label{eq:udpo}
\mathcal{L}_{\text{DPO-Uncert}}(\theta) = -\mathbb{E}_{(x, y^+, y^-, u)} u \cdot \log \sigma\left( \beta \cdot \Delta_{\theta, \text{ref}} \right).
\end{equation}


The strategy of our method is to introduce an uncertainty-weighted loss, which reframes preference comparisons from `hard labels' to `soft evidence' following \citet{zhou2024wpo}. This allows the learning process to prioritize preference with reliable answers. We apply this principle to both PPO and DPO-style optimization. In the PPO-style approach, the weights are used during reward model training to produce a more reliable reward signal $r_\phi(x,y)$. In the DPO-style approach, the weights are applied directly to the preference loss to modulate gradient updates. While the mechanisms differ, both methods ensure the final policy is shaped more significantly by high-confidence data for better alignment.

\begin{algorithm}[t]
\caption{Conformal Feedback Alignment (CFA)}
\label{alg:rluf}
\begin{algorithmic}[1]
\REQUIRE Initial policy $\pi_\theta$, reference policy $\pi_{\text{ref}}$
\FOR{each training iteration}
\STATE Collect preference pairs ${(x, y^+, y^-)}$ using AI or human feedback
\FOR{each pair $(x, y^+, y^-)$}
\STATE Estimate nonconformity scores $s(x, y^+), s(x, y^-)$ via CP
\STATE Compute set-wise uncertainty score $u(x, y^+, y^-)$ according to Eq.\eqref{eq:uncertainty}
\ENDFOR
\STATE Form augmented dataset $\mathcal{D}_{\text{uncert}} = {(x, y^+, y^-, u)}$
\IF{PPO-style}
\STATE Train reward model $r_\phi$ with Eq.\eqref{eq:uppo}
\STATE Fine-tune policy $\pi_\theta$ using PPO with $r_\phi$
\ELSE
\STATE Update policy $\pi_\theta$ with Eq.\eqref{eq:udpo}
\ENDIF
\ENDFOR
\end{algorithmic}
\end{algorithm}

\subsection{Algorithm Description}
We provide a summary of the procedure in Algorithm~\ref{alg:rluf}. This algorithm describes the complete pipeline of \ours, covering both the uncertainty estimation and the policy optimization stages.
Our algorithm begins by initializing the policy $\pi_\theta$ and reference $\pi_{\text{ref}}$, and calibrating nonconformity score thresholds. Each iteration involves two stages. First, we collect preference data $(x, y^+, y^-)$ and estimate an uncertainty score $u$ for each sample via CP (Section~\ref{sec:cp-uq}), yielding an uncertainty-augmented dataset $\mathcal{D}_{\text{uncert}}$. Second, we update the policy $\pi_\theta$ using this dataset. This can be done with uncertainty-aware alignment introduced in Section~\ref{sec:upl} with PPO-style or DPO-style training.
This algorithm allows the policy to focus on preference comparisons that are built from reliable answers, while reducing the influence of noisy or ambiguous feedback. By integrating CP into the learning loop, the method calibrates the learning signal dynamically and adapts to the uncertainty present in the data.

%

\begin{table*}[ht]
  \centering
  \resizebox{\textwidth}{!}{
  \begin{tabular}{l *{3}{c} *{3}{c} *{3}{c}}
    \toprule
    Dataset 
      & \multicolumn{3}{c}{Summarize} 
      & \multicolumn{3}{c}{Pairwise} 
      & \multicolumn{3}{c}{WebGPT} \\
    \cmidrule(lr){2-4} \cmidrule(lr){5-7} \cmidrule(lr){8-10}
    Models 
      & SFT & Base & \ours
      & SFT & Base & \ours 
      & SFT & Base & \ours \\
    \midrule
\multicolumn{10}{c}{PPO}   \\\midrule                               
Llama2-7B   & 63.91 & 65.88 & \textbf{67.39 (+1.51)} & 89.32 & 91.15 & \textbf{91.89 (+0.74)} & 69.54 & 72.25  & \textbf{72.78 (+0.53)}\\
Llama3.1-8B & 56.18 & 56.93 & \textbf{57.59 (+0.66)} & 79.22 & 81.38 & \textbf{82.25 (+0.87)} & 72.85 & 75.02 & \textbf{75.56 (+0.54)}  \\
Qwen2.5-7B  & 52.42 & 54.29 & \textbf{55.21 (+0.92)} & 88.25 & 90.17 & \textbf{90.65 (+0.48)} & 73.37  & 75.64  & \textbf{75.95 (+0.31)} \\
\midrule  
\multicolumn{10}{c}{DPO}                                                                                                             \\ \midrule   
Llama2-7B   & 63.91 & 65.68 & \textbf{67.30 (+1.62)} & 89.32 & 90.88 & \textbf{92.12 (+1.24)} & 69.54 & 71.68 & \textbf{72.25 (+0.57)} \\
Llama3.1-8B & 56.18 & 56.7  & \textbf{57.44 (+0.74)} & 79.22 & 81.55 & \textbf{82.90 (+1.35)} & 72.85 & 74.70 & \textbf{75.19 (+0.49)} \\
Qwen2.5-7B  & 52.42 & 53.85 & \textbf{55.01 (+1.16)} & 88.25 & 89.82 & \textbf{90.97 (+1.15)} & 73.37 & 75.91 & \textbf{76.42 (+0.51)} \\\bottomrule   
  \end{tabular}
  }
  \caption{Performance comparison of our methods with baselines across various datasets and models using LLM-as-a-Judge. The average scores are reported. The higher the score, the better the performance.  SFT is supervised fine-tuning only without any preference alignment. Base is the normal PPO or DPO alignment without uncertainty. The \textbf{best} result is shown. The numbers in parentheses indicate improvements over the base alignment method. \ours outperforms base methods across different models and datasets consistently. }
  \vspace{-2mm}
  \label{tab:main_results}
\end{table*}

\section{Experiments}
In this section, we mainly conduct experiments to answer the following research questions:\\
\noindent$\bullet$~\textbf{RQ1:} Does the proposed method outperform normal alignment or post-training methods overall?\\
\noindent$\bullet$~\textbf{RQ2:} How Does \ours perform under different scales of model parameters and data?\\
\noindent$\bullet$~\textbf{RQ3:} How does \ours perform against preference-level uncertainty-aware methods, and how is it affected by white-box versus black-box settings?

\subsection{Experimental Setup}
\label{sec:experiment_setup}

\paragraph{Models} For a comprehensive evaluation of the performance of our proposed method, we use in total three different open-source models as the pre-trained model. In detail, we use Llama2-7b~\cite{touvron2023llama} and Llama3.1-8b~\cite{grattafiori2024llama} to show that our method works for different versions of the popular open-source models. We also use Qwen series, including Qwen2.5-7B~\cite{yang2024qwen2} and Qwen3 series~\cite{yang2025qwen3} as the pre-trained model to show that our method could be generalized to different model architectures. For all models, we use a standard framework that first does a supervised fine-tuning and then does alignment.

\paragraph{Dataset}
In this paper, we use three datasets:

\noindent$\bullet$~Webgpt Comparisons (Webgpt) ~\cite{nakano2021webgpt}: A dataset of question-answering outputs from the WebGPT model, used to evaluate long-form answers grounded in web search results.

\noindent$\bullet$~Synthetic Instruct GPT-J Pairwise (Pairwise) ~\cite{alex2021online}: A general-purpose instruction following dataset for alignment via synthetic prompts and responses in a wide range of tasks.

\noindent$\bullet$~Summarize from Feedback (Summarize)~\cite{stiennon2020learning}: A summarization dataset targeting content compression, focusing on Reddit posts and learning from preference-based feedback.

For each dataset, our method will generate new answers using the question in the original dataset with conformal prediction for each model. Table~\ref{tab:data-distribution} summarizes the detailed sizes of the augmented dataset for each model to be evaluated.

\begin{table}[h]
\centering
\resizebox{0.48\textwidth}{!}{
\begin{tabular}{lccc}
\hline
       & \textbf{Webgpt} & \textbf{Pairwise} & \textbf{Summarize} \\
\hline
\textbf{Llama2-7B}  & 53199 & 68067  &   104614 \\
\textbf{Llama3.1-8B}  & 55465 & 62101  & 101191  \\
\textbf{Qwen2.5-7B}  & 52101 & 68029  &  103845 \\
\hline
\end{tabular}
}
\vspace{-3mm}
\caption{Augmented dataset size for each model.}
\label{tab:data-distribution}
\vspace{-3mm}
\end{table}

\paragraph{Evaluation} Following previous work~\cite{zhang2024prototypical}, all outputs from trained models are evaluated by LLM-as-a-judge, which we use GPT-4o~\cite{achiam2023gpt}. This methodology aligns with various studies~\cite{gilardi2023chatgpt,alizadeh2023open} highlighting the capabilities of LLMs to produce high-quality text
evaluation that aligns with or surpasses human. Specifically, four different dimensions: (1) \textit{Accuracy}, which assesses whether the content of the answer or summary correctly reflects the information and intent of the original prompt.
(2) \textit{Relevance}, which checks whether the answer or summary closely aligns with the subject of the prompt.
(3) \textit{Completeness}, which evaluates whether the response includes all essential points and details from the prompt.
(4) \textit{Expression}, which considers whether the language used in the answer or summary is clear and easy to understand.
Each criterion is scored up to 100 in increments of 5.
Due to page limits, we only report the average of these four scores in the main results. The evaluation prompts can be found in Figure~\ref{fig:prompt}.

\begin{figure*}[t]
\centering
\begin{minipage}[t]{0.32\linewidth}
  \centering
  \includegraphics[width=\linewidth]{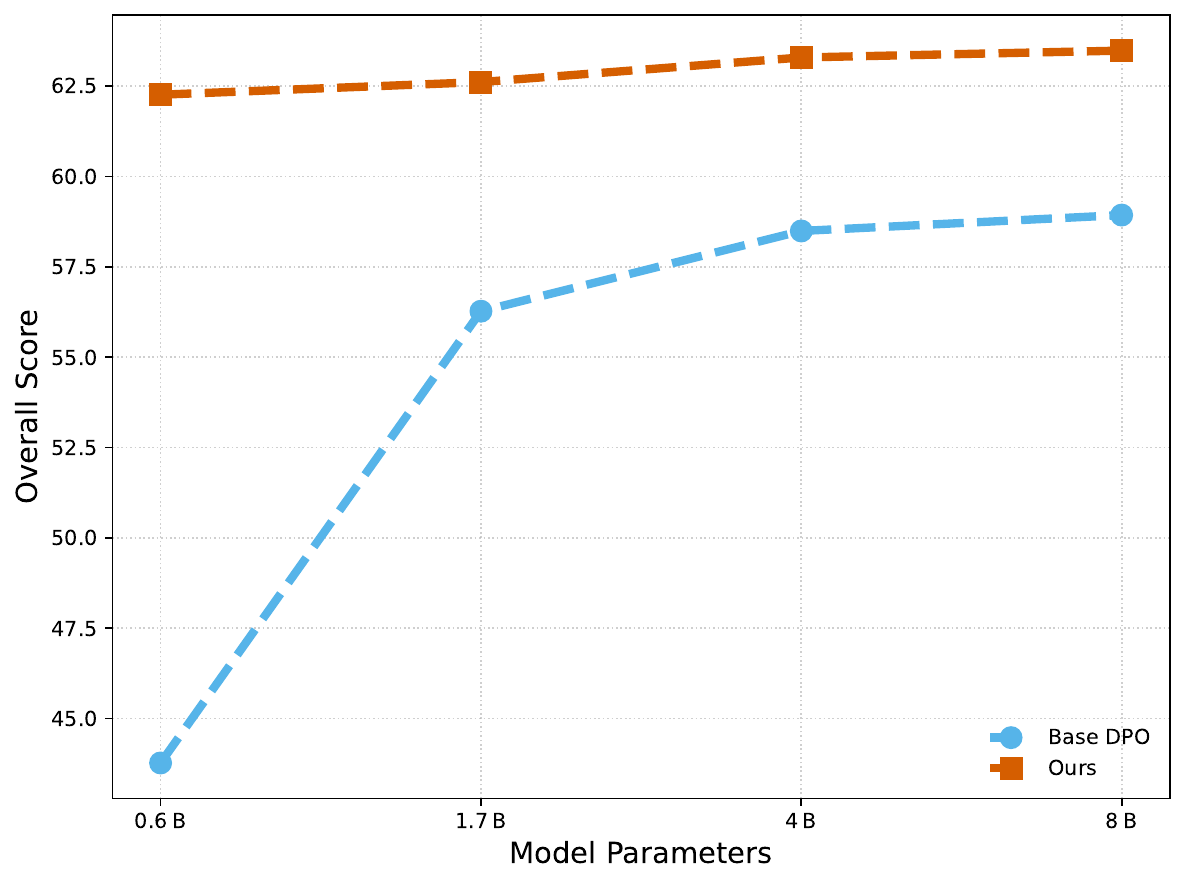}
  \captionof{figure}{Performance comparison on Llama2-7B and the Summarize dataset with different sizes of model parameters.}
  \label{fig:model_parameter}
\end{minipage}\hfill
\begin{minipage}[t]{0.32\linewidth}
  \centering
  \includegraphics[width=\linewidth]{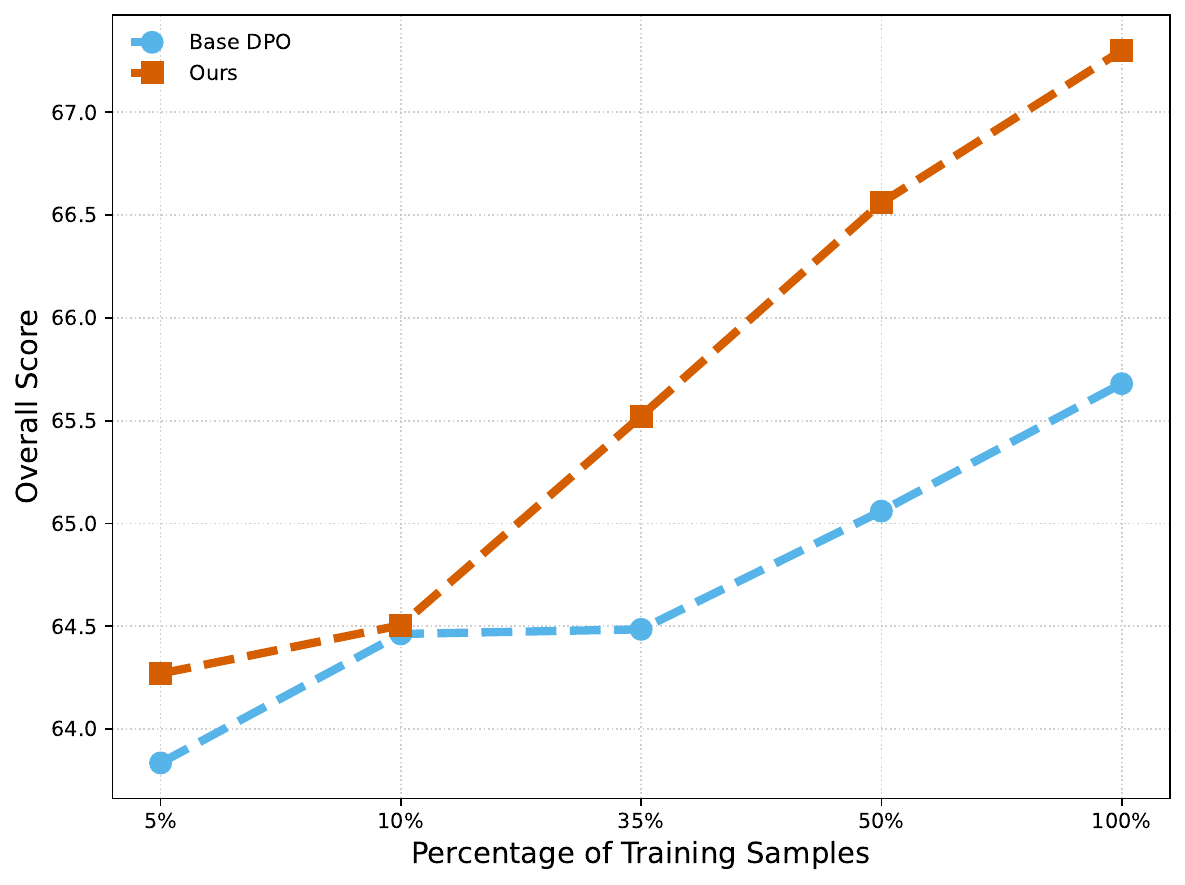}
  \captionof{figure}{Performance comparison on Llama2-7B and the Summarize dataset with different sizes of training samples.}
  \label{fig:training_samples}
\end{minipage}\hfill
\begin{minipage}[t]{0.32\linewidth}
  \centering
  \includegraphics[width=\linewidth]{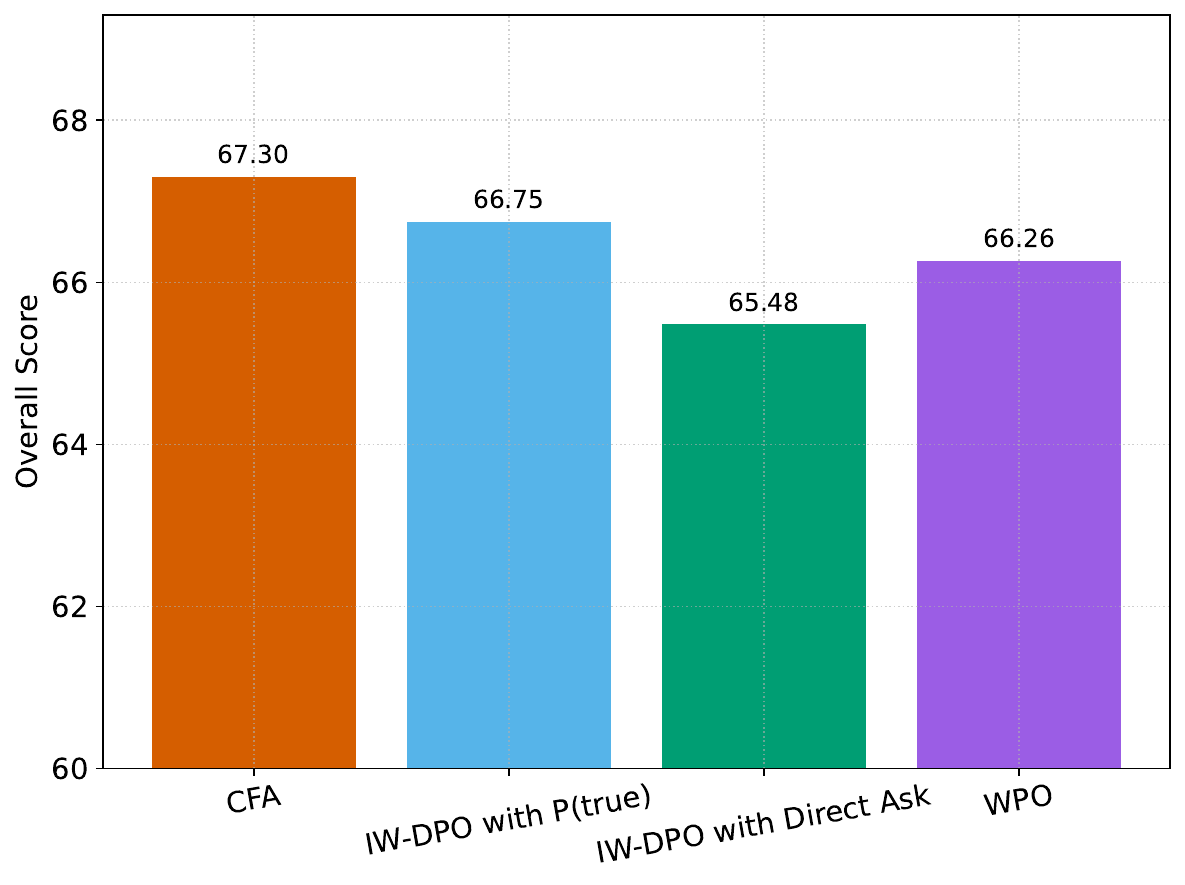}
  \captionof{figure}{Performance comparison on Llama2-7B and the Summarize dataset using \ours and other methods.}
  \label{fig:ptrue}
\end{minipage}
\end{figure*}

\paragraph{Implementations} Due to hardware constraints, for all experiments, our experiment applies a batch size of 1. We use the coverage $\alpha=0.8$ and $\alpha=0.5$ in the main results. We set the maximum generation length for both the CP process and test process after training to 1024. We use the AdamW optimizer~\cite{zhuang2022understanding} and use the learning rate $1e-6$. For black-box CP settings, we use a lower temperature = 0.15 since we need repeated answers. For white-box CP settings, we use a default temperature = 0.7 as in previous work~\cite{quach2023conformal}.  For all datasets, we regenerate the samples according to the instructions in the data and use GPT-4~\cite{achiam2023gpt} and Alpacafarm~\cite{dubois2023alpacafarm} to obtain AI feedback on preference. For the CP calibration set, we randomly sample 100 samples from each dataset as the calibration set and ensure there is no overlap between the calibration set and the training set. All of experiments on done using 4 Nvidia-A100 GPUs.\\

\begin{figure}[H]
\centering
\begin{tcolorbox}[title=Prompt for Evaluation]
\footnotesize  %

You are an expert evaluator. You are given an original input and an AI-generated response. Your task is to evaluate
the response based on four criteria: Accuracy, Relevance, Completeness, and
Expression. Each criterion should be scored from 0 to 100 in increments of 5.
Provide a brief justification for each score. Then, calculate the average of the
four scores and present it as the Overall Score.

\medskip
\textbf{Scoring Criteria:}
\begin{enumerate}[label=\textbf{\arabic*.}, itemsep=0.15em, leftmargin=1.0em]
  \item Accuracy (Acc): Does the response accurately reflect the content and intent of the original prompt?
  \item Relevance (Rel): Is the response closely aligned with the topic and requirements of the prompt?
  \item Completeness (Comp): Does the response address all essential aspects or key points in the prompt?
  \item Expression (Expr): Is the response clear, well-written, and easy to understand?
\end{enumerate}

\medskip
\textbf{Please \emph{only} return the four line‐scores and the Overall Score, in this exact format:}

\begin{verbatim}
**Accuracy (Acc):** [score]/10
**Relevance (Rel):** [score]/10
**Completeness (Comp):** [score]/10
**Expression (Expr):** [score]/10
**Overall Score:** [average]/10
\end{verbatim}
\end{tcolorbox}
\caption{Prompt for Evaluation in our Test Stage.}
\label{fig:prompt}
\end{figure}

\begin{figure*}[t!]
\centering
\includegraphics[width=0.8\textwidth]{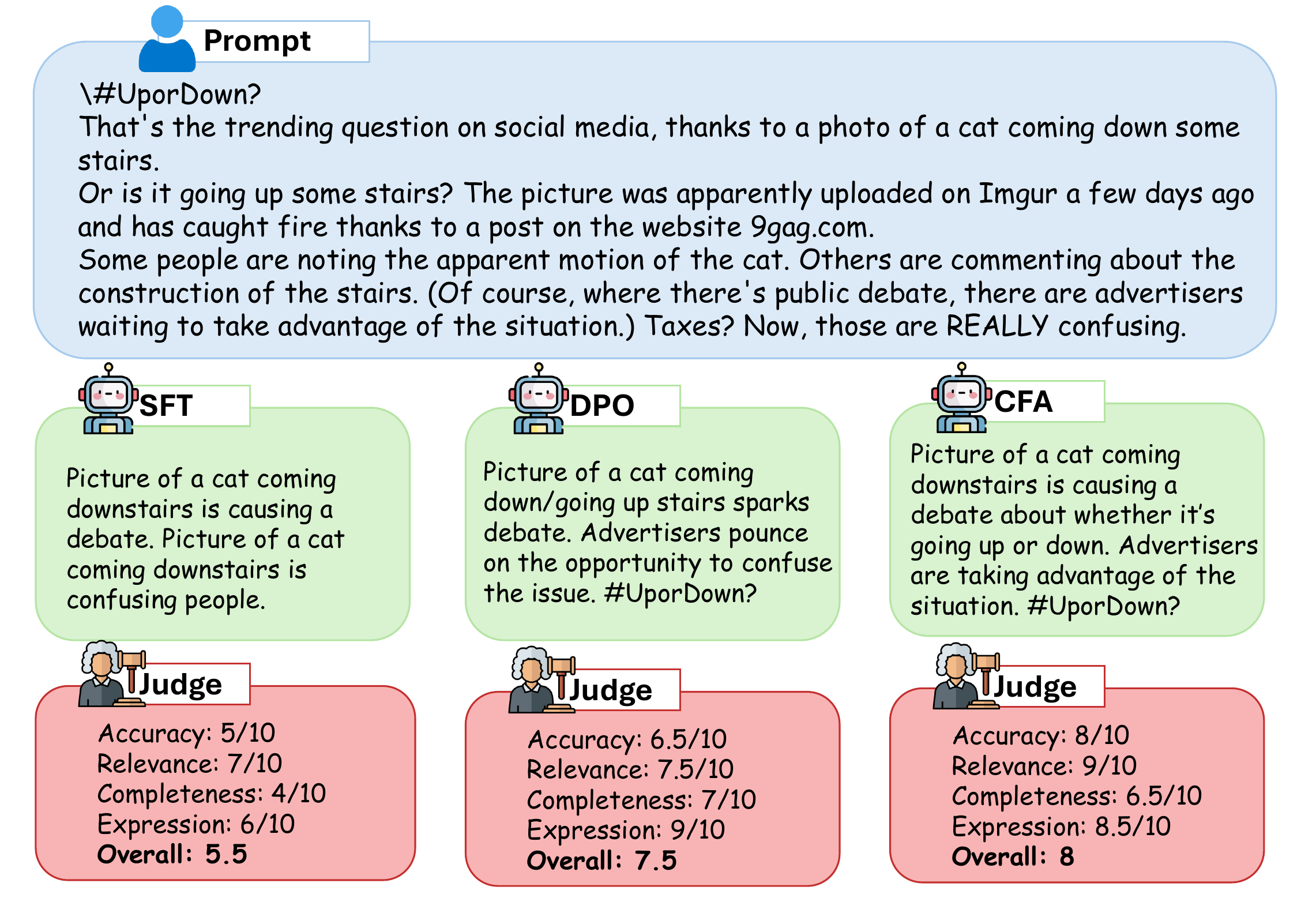}
\caption{A case study on the Summarization dataset compared with the summarization output from the model with SFT, the model with Base DPO, and the model with our methods. The output results and scores show a clear advantage of our method.}
\label{fig:sample}
\end{figure*}

\subsection{Comparison with Baselines (RQ1)}


To evaluate the effectiveness of our method, \ours, we compare it against several standard alignment baselines: Supervised Fine-tuning (SFT), PPO, and DPO. The evaluation is conducted on three distinct language models across three benchmark datasets. Following the experimental setup detailed in Section~\ref{sec:experiment_setup}, we use GPT-4o for automated evaluation. The main results are presented in Table~\ref{tab:main_results}, from which we draw the following observations:

\noindent$\bullet$~Our method consistently achieves superior performance across all evaluated scenarios. For instance, on the Llama2-7B model with the Summarize dataset, the improvement of \ours over the base model is +1.62, which is comparable to the +1.77 gain achieved by the base model over SFT. 

\noindent$\bullet$~Considering the performance gain using DPO and PPO, we observe that our method combined with DPO-style training yields greater improvements. One potential reason is that our method is used to train the reward model for PPO-style training. However, in PPO, the reward model is only one component, and the overall training process is more complex and harder to optimize effectively. \\

\vspace{-2mm}
\subsection{Scalability and Efficiency Analysis (RQ2)}
To evaluate the scalability and data efficiency of our method, we examine its performance across different model sizes and varying proportions of training data. This helps assess its practical applicability under diverse resource and data constraints. All experiments use black-box CP.

\paragraph{Influence of Different Model Parameters} To understand how different sizes of model parameters influence performance, we train Qwen3 series~\cite{yang2025qwen3}, which contains models from 0.6B to 8B, allowing for comprehensive analysis. In detail, we conduct \ours with DPO on four different sizes of models from the Qwen3 series and the Summarize dataset~\cite{stiennon2020learning}. In Figure~\ref{fig:model_parameter}, we show the overall score for four different models. The results show that \ours works stably for different sizes of models, and the performance gain is even larger for the smaller model, which shows the effectiveness of \ours.

\paragraph{Influence of Different Sizes of Training Samples} To understand how different sizes of training samples, we train Llama2-7B with $5\%$, $10\%$, $20\%$, $35\%$, $50\%$ of the data in the Summarize dataset. The results can be found in Figure~\ref{fig:training_samples}. We could see that for different proportions of the training data usage, \ours consistently outperforms the base DPO even when the data proportion is as low as 5\%. And with the increase in training samples, \ours shows a higher performance gain, which demonstrates the scalability of \ours. 


\subsection{Ablation Study and Sensitivity Analysis (RQ3)}

In this section, we are going to answer How does \ours perform against preference-level uncertainty-aware methods and how different settings of conformal prediction influence the performance. More experiments about the settings of conformal prediction can be found in the Appendix.

\paragraph{Performance for preference-level uncertainty-aware methods} To demonstrate the effectiveness of using answer-level reliability, we compare \ours with WPO~\cite{zhou2024wpo} as well as IW-DPO~\cite{lodkaewimportance} with the following uncertainty signals as the weight: \\
\noindent $\bullet$ \textit{Direct Ask}: When obtaining preferences of answers, directly ask LLMs to output the confidence of this preference. \\
\noindent $\bullet$ \textit{p(true)}~\cite{kadavath2022language}: When obtaining preferences of answers, using the output probability of the chosen preference as the confidence.
\textit{Entropy}

We train the Llama2-7B on the Summarize dataset with DPO, and the detailed comparison can be found at Figure~\ref{fig:ptrue}. The results show that \ours outperforms IW-DPO using different UQ methods and WPO. For IW-DPO using \textit{Direct Ask} as the weight, it even shows a worse performance than the base DPO, showing the unreliability of uncertainty from LLMs themselves.

\begin{figure}[t!]
\centering
\includegraphics[width=0.4\textwidth]{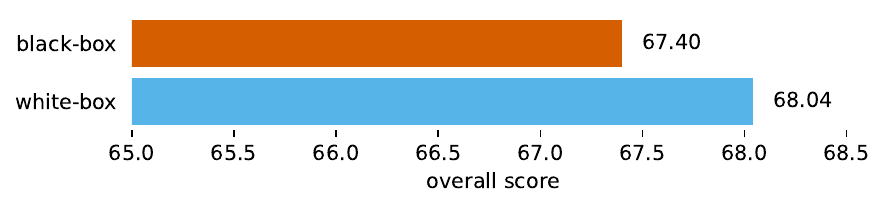}
\caption{The comparison using white-box and black-box CP on Llama2-7B and Summarize Dataset. The result shows that using white-box CP could lead to an even better performance.}
\vspace{-5mm}
\label{fig:white_box}
\end{figure}

\paragraph{White-box v.s. Black-box} 
Previously, we used black-box CP to ensure the generalization ability of \ours. Here, to see how white-box CP influences the performance, we show a performance comparison using white-box and black-box CP on Llama2-7B with the Summarize dataset. The results in Figure~\ref{fig:white_box} show that using white-box CP, which allows for generating diverse answers, could lead to an even better performance, showing the potential of \ours.

\noindent \textbf{Win Rate Comparisons} To better understand the win rates between the responses of models trained with \ours and normal DPO. Same with the main text, we are using GPT-4o to judge which answer is better. The result can be found in Table~\ref{tab:winrate}.  Results show a clear advantage over \ours.

\begin{table}[H]
\centering
\resizebox{0.48\textwidth}{!}{%
\begin{tabular}{c|ccc}
\toprule
WinRate &  Llama2 & Llama3.1  & Qwen2.5 \\
\midrule
\ours v.s Base\_DPO  & 56.18\%  & 64.33\%  & 57.61\%\\
\bottomrule
\end{tabular}%
}
\caption{The Win rates compared with \ours with Base DPO on three models on the Summarize dataset. The results clearly show that \ours has an advantage.}
\label{tab:winrate}
\end{table}

\begin{figure}[t!]
\centering
\includegraphics[width=0.45\textwidth]{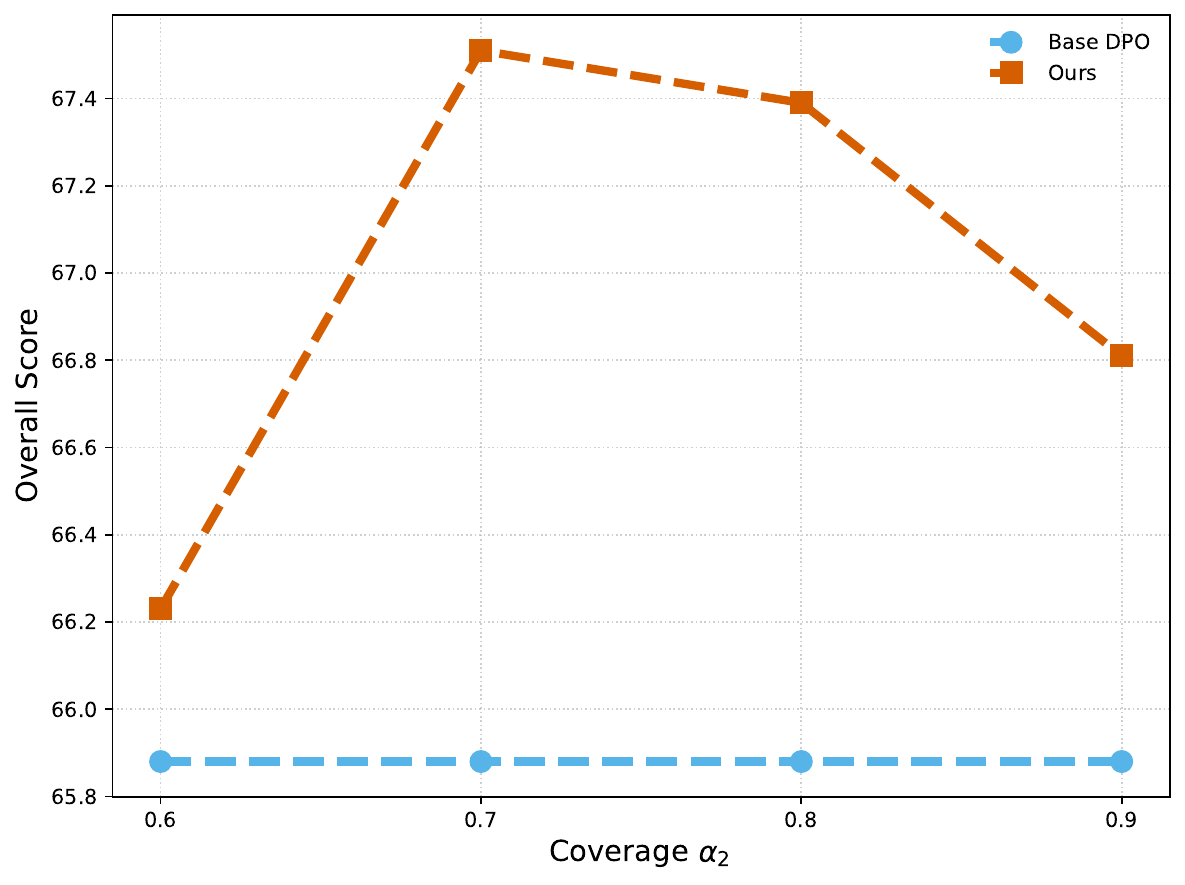}
\caption{The comparison using different coverage $\alpha_2$ in \ours on Llama2-7B and Summarize Dataset. The result shows there are some sweet points for the Coverage $\alpha_2$, but \ours consistently outperforms the base DPO.}
\label{fig:coverage}
\end{figure}

\noindent \textbf{Different Coverage in CP} In the main experiments, we are using coverage $\alpha_1 = 0.5$ and $\alpha_2=0.8$. One advantage of using Conformal Prediction is that users can customize the coverage in the CP process. Therefore, in this section, we change the different $\alpha_2$ to see how \ours performs. In detail, we change $\alpha_2$ from 0.6 to 0.9 with Llama2-7B and the Summarize Dataset. The results are shown in Figure~\ref{fig:coverage}. The results show that using a coverage of 0.7-0.8 can have the best overall performance. When the $\alpha_2$ is 0.6, we get the worst performance because $\alpha_2$ is too close to $\alpha_1 = 0.5$, resulting in the reduction of the effect of weights in \ours. However, \ours consistently outperforms the base DPO, showing \ours is better for alignment of LLMs.

\subsection{Case Analysis}

To better understand the effect of uncertainty-aware preference modeling, we examine a representative example from the Summarize dataset in Figure~\ref{fig:sample}. The prompt involves summarizing an optical illusion debate featuring a photo of a cat. From the case, we can see that the SFT model's output is highly repetitive and uninformative, a fact reflected in its low overall score of 5.5 out of 10, which demonstrates why we need reinforcement learning.

The Base DPO model attempts to solve the problem from SFT model by incorporating more detail but misinterprets a critical part of the prompt. Specifically, it states that “advertisers pounce on the opportunity to confuse the issue,” incorrectly implying that advertisers are actively creating confusion, which is wrong.  This type of subtle misinterpretation is a key problem when a model learns from all feedback indiscriminately. It treats all feedback as equally valid, which prevents it from capturing the true, nuanced relationship described in the source text.

Our uncertainty-aware method excels because it corrects this specific failure. It achieved the highest scores for accuracy and relevance, 8 and 9, respectively, because it learns to disregard such noisy, misleading signals. By focusing on high-confidence feedback, it develops a more precise understanding, correctly identifying that advertisers are "taking advantage of the situation." This demonstrates its ability to filter out ambiguity and generate a summary that is not only fluent but also factually accurate and contextually relevant.

\section{Conclusion}
In this paper, we presented a new framework for Conformal Feedback Alignment (CFA), which improves the alignment of large language models by using the reliability of responses. Our method provides both reward-model-based and reward-free training paradigms under white-box and black-box settings. Empirical results show that incorporating confidence and reliability from answers into preference learning not only enhances alignment quality but also scales well with limited data and smaller models. Overall, this work highlights the importance of the reliability of answers in alignment pipelines and provides a practical, supported solution. Future research may explore extending this framework to multimodal feedback or human-in-the-loop calibration under real-world constraints.

\section*{Acknowledgment}
The work was partially supported by NSF award \#2442477. We thank Amazon Research Awards, Cisco Research Awards, Google, and OpenAI for providing us with API credits. The authors acknowledge Research Computing at Arizona State University for providing computing resources. The views and conclusions in this paper are those of the authors and should not be interpreted as representing any funding agencies.

\section*{Limitation}
While Conformal Feedback Alignment (CFA) demonstrates promising improvements in robustness for LLM alignment, several limitations remain. First, CFA relies on the quality of conformal prediction calibration. When the calibration set is small, the resulting reliability estimates may be inaccurate. Second, the current experiments are conducted on text-based models and dataset, lacking the exploration for the multi-modal large language model. Finally, our approach does not combine the answer reliability and preference uncertainty. Future work should investigate how to jointly model them to further enhance alignment performance. We use LLM for grammar check only.

\bibliography{custom}

@article{touvron2023llama,
  title={Llama 2: Open foundation and fine-tuned chat models},
  author={Touvron, Hugo and Martin, Louis and Stone, Kevin and Albert, Peter and Almahairi, Amjad and Babaei, Yasmine and Bashlykov, Nikolay and Batra, Soumya and Bhargava, Prajjwal and Bhosale, Shruti and others},
  journal={arXiv preprint arXiv:2307.09288},
  year={2023}
}

@article{grattafiori2024llama,
  title={The llama 3 herd of models},
  author={Grattafiori, Aaron and Dubey, Abhimanyu and Jauhri, Abhinav and Pandey, Abhinav and Kadian, Abhishek and Al-Dahle, Ahmad and Letman, Aiesha and Mathur, Akhil and Schelten, Alan and Vaughan, Alex and others},
  journal={arXiv preprint arXiv:2407.21783},
  year={2024}
}

@article{alizadeh2023open,
  title={Open-source large language models outperform crowd workers and approach ChatGPT in text-annotation tasks},
  author={Alizadeh, Meysam and Kubli, Ma{\"e}l and Samei, Zeynab and Dehghani, Shirin and Bermeo, Juan Diego and Korobeynikova, Maria and Gilardi, Fabrizio},
  journal={arXiv preprint arXiv:2307.02179},
  volume={101},
  year={2023},
  publisher={Technical Report}
}

@inproceedings{liu2025uncertainty,
author = {Liu, Xiaoou and Chen, Tiejin and Da, Longchao and Chen, Chacha and Lin, Zhen and Wei, Hua},
title = {Uncertainty Quantification and Confidence Calibration in Large Language Models: A Survey},
year = {2025},
isbn = {9798400714542},
publisher = {Association for Computing Machinery},
address = {New York, NY, USA},
url = {https://doi.org/10.1145/3711896.3736569},
doi = {10.1145/3711896.3736569},
booktitle = {Proceedings of the 31st ACM SIGKDD Conference on Knowledge Discovery and Data Mining V.2},
pages = {6107–6117},
numpages = {11},
location = {Toronto ON, Canada},
series = {KDD '25}
}

@article{da2024open,
  title={Open-ti: Open traffic intelligence with augmented language model},
  author={Da, Longchao and Liou, Kuanru and Chen, Tiejin and Zhou, Xuesong and Luo, Xiangyong and Yang, Yezhou and Wei, Hua},
  journal={International Journal of Machine Learning and Cybernetics},
  volume={15},
  number={10},
  pages={4761--4786},
  year={2024},
  publisher={Springer Berlin Heidelberg Berlin/Heidelberg}
}

@inproceedings{yao2025comal,
  title={CoMAL: Collaborative Multi-Agent Large Language Models for Mixed-Autonomy Traffic},
  author={Yao, Huaiyuan and Da, Longchao and Nandam, Vishnu and Turnau, Justin and Liu, Zhiwei and Pang, Linsey and Wei, Hua},
  booktitle={SDM 2025},
  year={2025}
}

@article{da2025survey,
  title={A survey of sim-to-real methods in rl: Progress, prospects and challenges with foundation models},
  author={Da, Longchao and Turnau, Justin and Kutralingam, Thirulogasankar Pranav and Velasquez, Alvaro and Shakarian, Paulo and Wei, Hua},
  journal={arXiv preprint arXiv:2502.13187},
  year={2025}
}

@inproceedings{da2025ge,
  title={GE-Chat: A Graph Enhanced RAG Framework for Evidential Response Generation of LLMs},
  author={Da, Longchao and Shah, Parth Mitesh and Liou, Kuan-Ru and Zhang, Jiaxing and Wei, Hua},
  booktitle={IJCAI 2025},
  year={2025}
}

@incollection{da2025flans,
  title={FlanS: A Foundation Model for Free-Form Language-based Segmentation in Medical Images},
  author={Da, Longchao and Wang, Rui and Xu, Xiaojian and Bhatia, Parminder and Kass-Hout, Taha and Wei, Hua and Xiao, Cao},
  booktitle={KDD 2025},
  pages={404--414},
  year={2025}
}

@inproceedings{da2025deepshade,
  title={Deepshade: Enable shade simulation by text-conditioned image generation},
  author={Da, Longchao and Liu, Xiangrui and Shivakoti, Mithun and Kutralingam, Thirulogasankar Pranav and Yang, Yezhou and Wei, Hua},
  booktitle={IJCAI 2025},
  pages={9610--9618},
  year={2025}
}

@inproceedings{yao2025instructional,
  title={Instructional agents: Llm agents on automated course material generation for teaching faculties},
  author={Yao, Huaiyuan and Xu, Wanpeng and Turnau, Justin and Kellam, Nadia and Wei, Hua},
  booktitle={EACL'26 Main Conference},
  year={2025},
  organization={19th Conference of the European Chapter of the Association for Computational~…}
}

@incollection{satheesh2025cmalc,
  title={cMALC-D: Contextual Multi-Agent LLM-Guided Curriculum Learning with Diversity-Based Context Blending},
  author={Satheesh, Anirudh and Powell, Keenan and Wei, Hua},
  booktitle={CIKM 2025},
  pages={5213--5217},
  year={2025}
}

@inproceedings{chen2025privacy,
  title={Privacy-preserving Fine-tuning of Large Language Models through Flatness},
  author={Chen, Tiejin and Da, Longchao and Zhou, Huixue and Li, Pingzhi and Zhou, Kaixiong and Chen, Tianlong and Wei, Hua},
  booktitle={SDM 2025},
  year={2025}
}

@inproceedings{chen2025protecting,
  title={Protecting Privacy against Membership Inference Attack with LLM Fine-tuning through Flatness},
  author={Chen, Tiejin and Da, Longchao and Zhou, Huixue and Li, Pingzhi and Zhou, Kaixiong and Chen, Tianlong and Wei, Hua},
  booktitle={SDM 2025},
  pages={386--397},
  year={2025},
  organization={Society for Industrial and Applied Mathematics}
}

@article{liu2025mcqa,
  title={MCQA-Eval: Efficient Confidence Evaluation in NLG with Gold-Standard Correctness Labels},
  author={Liu, Xiaoou and Lin, Zhen and Da, Longchao and Chen, Chacha and Trivedi, Shubhendu and Wei, Hua},
  journal={arXiv preprint arXiv:2502.14268},
  year={2025}
}

@article{da2024llm,
  title={Llm uncertainty quantification through directional entailment graph and claim level response augmentation},
  author={Da, Longchao and Chen, Tiejin and Cheng, Lu and Wei, Hua},
  journal={arXiv preprint arXiv:2407.00994},
  year={2024}
}

@article{gilardi2023chatgpt,
  title={ChatGPT outperforms crowd workers for text-annotation tasks},
  author={Gilardi, Fabrizio and Alizadeh, Meysam and Kubli, Ma{\"e}l},
  journal={Proceedings of the National Academy of Sciences},
  volume={120},
  number={30},
  pages={e2305016120},
  year={2023},
  publisher={National Academy of Sciences}
}

@article{yang2024qwen2,
  title={Qwen2. 5 technical report},
  author={Yang, An and Yang, Baosong and Zhang, Beichen and Hui, Binyuan and Zheng, Bo and Yu, Bowen and Li, Chengyuan and Liu, Dayiheng and Huang, Fei and Wei, Haoran and others},
  journal={arXiv preprint arXiv:2412.15115},
  year={2024}
}

@article{achiam2023gpt,
  title={Gpt-4 technical report},
  author={Achiam, Josh and Adler, Steven and Agarwal, Sandhini and Ahmad, Lama and Akkaya, Ilge and Aleman, Florencia Leoni and Almeida, Diogo and Altenschmidt, Janko and Altman, Sam and Anadkat, Shyamal and others},
  journal={arXiv preprint arXiv:2303.08774},
  year={2023}
}

@article{christiano2017deep,
  title={Deep reinforcement learning from human preferences},
  author={Christiano, Paul F and Leike, Jan and Brown, Tom and Martic, Miljan and Legg, Shane and Amodei, Dario},
  journal={Advances in neural information processing systems},
  volume={30},
  year={2017}
}

@article{ouyang2022training,
  title={Training language models to follow instructions with human feedback},
  author={Ouyang, Long and Wu, Jeffrey and Jiang, Xu and Almeida, Diogo and Wainwright, Carroll and Mishkin, Pamela and Zhang, Chong and Agarwal, Sandhini and Slama, Katarina and Ray, Alex and others},
  journal={Advances in neural information processing systems},
  volume={35},
  pages={27730--27744},
  year={2022}
}

@article{schulman2017proximal,
  title={Proximal policy optimization algorithms},
  author={Schulman, John and Wolski, Filip and Dhariwal, Prafulla and Radford, Alec and Klimov, Oleg},
  journal={arXiv preprint arXiv:1707.06347},
  year={2017}
}

@article{hu2025reinforce++,
  title={Reinforce++: A simple and efficient approach for aligning large language models},
  author={Hu, Jian},
  journal={arXiv preprint arXiv:2501.03262},
  year={2025}
}

@article{rafailov2023direct,
  title={Direct preference optimization: Your language model is secretly a reward model},
  author={Rafailov, Rafael and Sharma, Archit and Mitchell, Eric and Manning, Christopher D and Ermon, Stefano and Finn, Chelsea},
  journal={Advances in Neural Information Processing Systems},
  volume={36},
  pages={53728--53741},
  year={2023}
}

@inproceedings{azar2024general,
  title={A general theoretical paradigm to understand learning from human preferences},
  author={Azar, Mohammad Gheshlaghi and Guo, Zhaohan Daniel and Piot, Bilal and Munos, Remi and Rowland, Mark and Valko, Michal and Calandriello, Daniele},
  booktitle={International Conference on Artificial Intelligence and Statistics},
  pages={4447--4455},
  year={2024},
  organization={PMLR}
}

@article{wu2024alpha,
  title={$alpha$-DPO: Adaptive Reward Margin is What Direct Preference Optimization Needs},
  author={Wu, Junkang and Wang, Xue and Yang, Zhengyi and Wu, Jiancan and Gao, Jinyang and Ding, Bolin and Wang, Xiang and He, Xiangnan},
  journal={arXiv preprint arXiv:2410.10148},
  year={2024}
}

@article{xu2024contrastive,
  title={Contrastive preference optimization: Pushing the boundaries of llm performance in machine translation},
  author={Xu, Haoran and Sharaf, Amr and Chen, Yunmo and Tan, Weiting and Shen, Lingfeng and Van Durme, Benjamin and Murray, Kenton and Kim, Young Jin},
  journal={arXiv preprint arXiv:2401.08417},
  year={2024}
}

@article{saeidi2024triple,
  title={Triple preference optimization: Achieving better alignment with less data in a single step optimization},
  author={Saeidi, Amir and Verma, Shivanshu and RRV, Aswin and Baral, Chitta},
  journal={arXiv preprint arXiv:2405.16681},
  year={2024}
}

@article{bai2022constitutional,
  title={Constitutional ai: Harmlessness from ai feedback},
  author={Bai, Yuntao and Kadavath, Saurav and Kundu, Sandipan and Askell, Amanda and Kernion, Jackson and Jones, Andy and Chen, Anna and Goldie, Anna and Mirhoseini, Azalia and McKinnon, Cameron and others},
  journal={arXiv preprint arXiv:2212.08073},
  year={2022}
}

@article{lee2023rlaif,
  title={Rlaif vs. rlhf: Scaling reinforcement learning from human feedback with ai feedback},
  author={Lee, Harrison and Phatale, Samrat and Mansoor, Hassan and Mesnard, Thomas and Ferret, Johan and Lu, Kellie and Bishop, Colton and Hall, Ethan and Carbune, Victor and Rastogi, Abhinav and others},
  journal={arXiv preprint arXiv:2309.00267},
  year={2023}
}

@article{li2024hrlaif,
  title={Hrlaif: Improvements in helpfulness and harmlessness in open-domain reinforcement learning from ai feedback},
  author={Li, Ang and Xiao, Qiugen and Cao, Peng and Tang, Jian and Yuan, Yi and Zhao, Zijie and Chen, Xiaoyuan and Zhang, Liang and Li, Xiangyang and Yang, Kaitong and others},
  journal={arXiv preprint arXiv:2403.08309},
  year={2024}
}

@article{williams2024multi,
  title={Multi-objective Reinforcement learning from AI Feedback},
  author={Williams, Marcus},
  journal={arXiv preprint arXiv:2406.07295},
  year={2024}
}

@article{yu2024rlaif,
  title={Rlaif-v: Aligning mllms through open-source ai feedback for super gpt-4v trustworthiness},
  author={Yu, Tianyu and Zhang, Haoye and Yao, Yuan and Dang, Yunkai and Chen, Da and Lu, Xiaoman and Cui, Ganqu and He, Taiwen and Liu, Zhiyuan and Chua, Tat-Seng and others},
  journal={arXiv preprint arXiv:2405.17220},
  year={2024}
}

@article{ethayarajh2024kto,
  title={Kto: Model alignment as prospect theoretic optimization},
  author={Ethayarajh, Kawin and Xu, Winnie and Muennighoff, Niklas and Jurafsky, Dan and Kiela, Douwe},
  journal={arXiv preprint arXiv:2402.01306},
  year={2024}
}

@article{anil2023palm,
  title={Palm 2 technical report},
  author={Anil, Rohan and Dai, Andrew M and Firat, Orhan and Johnson, Melvin and Lepikhin, Dmitry and Passos, Alexandre and Shakeri, Siamak and Taropa, Emanuel and Bailey, Paige and Chen, Zhifeng and others},
  journal={arXiv preprint arXiv:2305.10403},
  year={2023}
}

@inproceedings{da2024prompt,
  title={Prompt to transfer: Sim-to-real transfer for traffic signal control with prompt learning},
  author={Da, Longchao and Gao, Minquan and Mei, Hao and Wei, Hua},
  booktitle={Proceedings of the AAAI Conference on Artificial Intelligence},
  volume={38},
  number={1},
  pages={82--90},
  year={2024}
}

@article{lin2023generating,
  title={Generating with confidence: Uncertainty quantification for black-box large language models},
  author={Lin, Zhen and Trivedi, Shubhendu and Sun, Jimeng},
  journal={arXiv preprint arXiv:2305.19187},
  year={2023}
}

@article{chen2025uncertainty,
  title={Uncertainty Quantification of Large Language Models through Multi-Dimensional Responses},
  author={Chen, Tiejin and Liu, Xiaoou and Da, Longchao and Chen, Jia and Papalexakis, Vagelis and Wei, Hua},
  journal={arXiv preprint arXiv:2502.16820},
  year={2025}
}

@article{quach2023conformal,
  title={Conformal language modeling},
  author={Quach, Victor and Fisch, Adam and Schuster, Tal and Yala, Adam and Sohn, Jae Ho and Jaakkola, Tommi S and Barzilay, Regina},
  journal={arXiv preprint arXiv:2306.10193},
  year={2023}
}

@article{su2024api,
  title={Api is enough: Conformal prediction for large language models without logit-access},
  author={Su, Jiayuan and Luo, Jing and Wang, Hongwei and Cheng, Lu},
  journal={arXiv preprint arXiv:2403.01216},
  year={2024}
}

@article{kadavath2022language,
  title={Language models (mostly) know what they know},
  author={Kadavath, Saurav and Conerly, Tom and Askell, Amanda and Henighan, Tom and Drain, Dawn and Perez, Ethan and Schiefer, Nicholas and Hatfield-Dodds, Zac and DasSarma, Nova and Tran-Johnson, Eli and others},
  journal={arXiv preprint arXiv:2207.05221},
  year={2022}
}

@article{band2024linguistic,
  title={Linguistic calibration of long-form generations},
  author={Band, Neil and Li, Xuechen and Ma, Tengyu and Hashimoto, Tatsunori},
  journal={arXiv preprint arXiv:2404.00474},
  year={2024}
}

@article{nakano2021webgpt,
  title={Webgpt: Browser-assisted question-answering with human feedback},
  author={Nakano, Reiichiro and Hilton, Jacob and Balaji, Suchir and Wu, Jeff and Ouyang, Long and Kim, Christina and Hesse, Christopher and Jain, Shantanu and Kosaraju, Vineet and Saunders, William and others},
  journal={arXiv preprint arXiv:2112.09332},
  year={2021}
}

@inproceedings{alex2021online,
  title={The Online Pivot: Lessons Learned from Teaching a Text and Data Mining Course in Lockdown, Enhancing online Teaching with Pair Programming and Digital Badges},
  author={Alex, Beatrice and Llewellyn, Clare and Orzechowski, Pawel and Boutchkova, Maria},
  booktitle={Proceedings of the Fifth Workshop on Teaching NLP},
  pages={138--148},
  year={2021}
}

@article{stiennon2020learning,
  title={Learning to summarize with human feedback},
  author={Stiennon, Nisan and Ouyang, Long and Wu, Jeffrey and Ziegler, Daniel and Lowe, Ryan and Voss, Chelsea and Radford, Alec and Amodei, Dario and Christiano, Paul F},
  journal={Advances in neural information processing systems},
  volume={33},
  pages={3008--3021},
  year={2020}
}

@inproceedings{zhang2024prototypical,
  title={Prototypical Reward Network for Data-Efficient RLHF},
  author={Zhang, Jinghan and Wang, Xiting and Jin, Yiqiao and Chen, Changyu and Zhang, Xinhao and Liu, Kunpeng},
  booktitle={Proceedings of the 62nd Annual Meeting of the Association for Computational Linguistics (Volume 1: Long Papers)},
  pages={13871--13884},
  year={2024}
}

@article{zhuang2022understanding,
  title={Understanding adamw through proximal methods and scale-freeness},
  author={Zhuang, Zhenxun and Liu, Mingrui and Cutkosky, Ashok and Orabona, Francesco},
  journal={arXiv preprint arXiv:2202.00089},
  year={2022}
}

@article{dubois2023alpacafarm,
  title={Alpacafarm: A simulation framework for methods that learn from human feedback},
  author={Dubois, Yann and Li, Chen Xuechen and Taori, Rohan and Zhang, Tianyi and Gulrajani, Ishaan and Ba, Jimmy and Guestrin, Carlos and Liang, Percy S and Hashimoto, Tatsunori B},
  journal={Advances in Neural Information Processing Systems},
  volume={36},
  pages={30039--30069},
  year={2023}
}

@article{yang2025qwen3,
  title={Qwen3 technical report},
  author={Yang, An and Li, Anfeng and Yang, Baosong and Zhang, Beichen and Hui, Binyuan and Zheng, Bo and Yu, Bowen and Gao, Chang and Huang, Chengen and Lv, Chenxu and others},
  journal={arXiv preprint arXiv:2505.09388},
  year={2025}
}

@article{shafer2008tutorial,
  title={A tutorial on conformal prediction.},
  author={Shafer, Glenn and Vovk, Vladimir},
  journal={Journal of Machine Learning Research},
  volume={9},
  number={3},
  year={2008}
}

@article{banerjee2024towards,
  title={Towards reliable alignment: Uncertainty-aware rlhf},
  author={Banerjee, Debangshu and Gopalan, Aditya},
  journal={arXiv preprint arXiv:2410.23726},
  year={2024}
}

@article{wang2024uncertainty,
  title={Uncertainty aware learning for language model alignment},
  author={Wang, Yikun and Zheng, Rui and Ding, Liang and Zhang, Qi and Lin, Dahua and Tao, Dacheng},
  journal={arXiv preprint arXiv:2406.04854},
  year={2024}
}

@article{ye2025robust,
  title={Robust reinforcement learning from human feedback for large language models fine-tuning},
  author={Ye, Kai and Zhou, Hongyi and Zhu, Jin and Quinzan, Francesco and Shi, Chengchun},
  journal={arXiv preprint arXiv:2504.03784},
  year={2025}
}

@inproceedings{xu2024uncertainty,
  title={Uncertainty-aware preference alignment in reinforcement learning from human feedback},
  author={Xu, Sheng and Yue, Bo and Zha, Hongyuan and Liu, Guiliang},
  booktitle={ICML 2024 Workshop on Models of Human Feedback for AI Alignment},
  year={2024}
}

@article{zhou2024wpo,
  title={Wpo: Enhancing rlhf with weighted preference optimization},
  author={Zhou, Wenxuan and Agrawal, Ravi and Zhang, Shujian and Indurthi, Sathish Reddy and Zhao, Sanqiang and Song, Kaiqiang and Xu, Silei and Zhu, Chenguang},
  journal={arXiv preprint arXiv:2406.11827},
  year={2024}
}

@article{lodkaewimportance,
  title={Importance Weighting for Aligning Language Models under Deployment Distribution Shift},
  author={Lodkaew, Thanawat and Fang, Tongtong and Ishida, Takashi and Sugiyama, Masashi},
  journal={Transactions on Machine Learning Research}
}


\end{document}